\documentclass[10pt,twocolumn,letterpaper]{article}

\usepackage{cvpr}              % To produce the CAMERA-READY version
\usepackage{graphicx}
\usepackage{amsmath}
\usepackage{amssymb}
\usepackage{booktabs}
\usepackage{dsfont}
\usepackage{amsfonts,amssymb}
\usepackage{bbm}
\usepackage{graphics}
\usepackage{tikz}
\usepackage{multirow, graphicx}
\usepackage{colortbl}
\usepackage{bbding}
\usepackage{tabularx}
\usepackage[marginal]{footmisc}
\renewcommand{\thefootnote}{}
\definecolor{cvprblue}{rgb}{0.21,0.49,0.74}
\usepackage[pagebackref,breaklinks,colorlinks,citecolor=cvprblue]{hyperref}

\def\paperID{3060} % *** Enter the Paper ID here
\def\confName{CVPR}
\def\confYear{2024}

\title{Hunting Attributes: Context Prototype-Aware Learning for \\ Weakly Supervised Semantic Segmentation}

\author{
Feilong Tang$^{1,2}$\footnotemark[1] \quad Zhongxing Xu$^{3}$\footnotemark[1] \quad Zhaojun Qu$^{4}$ \quad Wei Feng$^{1,2}$ \\ Xingjian Jiang$^{5}$ \quad Zongyuan Ge$^{1,2}$\footnotemark[2] \\ \\
\textsuperscript{\rm 1}{AIM Lab, Faculty of IT, Monash University}\quad
\textsuperscript{\rm 2}{Faculty of IT, Monash University} \\ 
\textsuperscript{\rm 3}{Weill Cornell Medicine, Cornell University}\quad
\textsuperscript{\rm 4}{Xi’an Jiaotong-Liverpool University}\\
\textsuperscript{\rm 5}{Ann Arbor, University of Michigan} \\
{\small \texttt{\{feilong.tang,zongyuan.ge\}@monash.edu}}
}
\begin{document}
\maketitle
\renewcommand{\thefootnote}{\fnsymbol{footnote}}
% \footnotetext[1]{These authors contributed equally to this work.}
\footnotetext[1]{The first two authors contribute equally to this work.}
\footnotetext[2]{Corresponding author: Zongyuan Ge}

%%%%%%%%% ABSTRACT
\vspace{-1cm}
\begin{abstract}
Recent weakly supervised semantic segmentation (WSSS) methods strive to incorporate contextual knowledge to improve the completeness of class activation maps (CAM). In this work, we argue that the knowledge bias between instances and contexts affects the capability of the prototype to sufficiently understand instance semantics. Inspired by prototype learning theory, we propose leveraging prototype awareness to capture diverse and fine-grained feature attributes of instances. The hypothesis is that contextual prototypes might erroneously activate similar and frequently co-occurring object categories due to this knowledge bias. Therefore, we propose to enhance the prototype representation ability by mitigating the bias to better capture spatial coverage in semantic object regions. With this goal, we present a Context Prototype-Aware Learning (CPAL) strategy, which leverages semantic context to enrich instance comprehension. The core of this method is to accurately capture intra-class variations in object features through context-aware prototypes, facilitating the adaptation to the semantic attributes of various instances. We design feature distribution alignment to optimize prototype awareness, aligning instance feature distributions with dense features. In addition, a unified training framework is proposed to combine label-guided classification supervision and prototypes-guided self-supervision. Experimental results on PASCAL VOC 2012 and MS COCO 2014 show that CPAL significantly improves off-the-shelf methods and achieves state-of-the-art performance. The project is available at 
\href{https://github.com/Barrett-python/CPAL}{https://github.com/Barrett-python/CPAL.} 

\end{abstract}

\vspace{-0.5cm}
%%%%%%%%% Introduction
\section{Introduction}
\label{sec:intro}
\quad Semantic segmentation serves as a fundamental task in the field of computer vision. Weakly Supervised Semantic Segmentation (WSSS) has become a popular approach in the community, learning from weak labels such as image-level labels \cite{kolesnikov2016seed, lee2021anti}, scribbles \cite{lin2016scribblesup,vernaza2017learning}, or bounding boxes \cite{dai2015boxsup,lee2021bbam,song2019box}, instead of pixel-level annotations. Most WSSS approaches utilize Class Activation Mapping (CAM) \cite{zhou2016learning} to provide localization cues for target objects, thereby mapping visual concepts to pixel regions.
\begin{figure}[t]
\centering
\begin{tabular}{cc}
\includegraphics[width=0.45\textwidth]{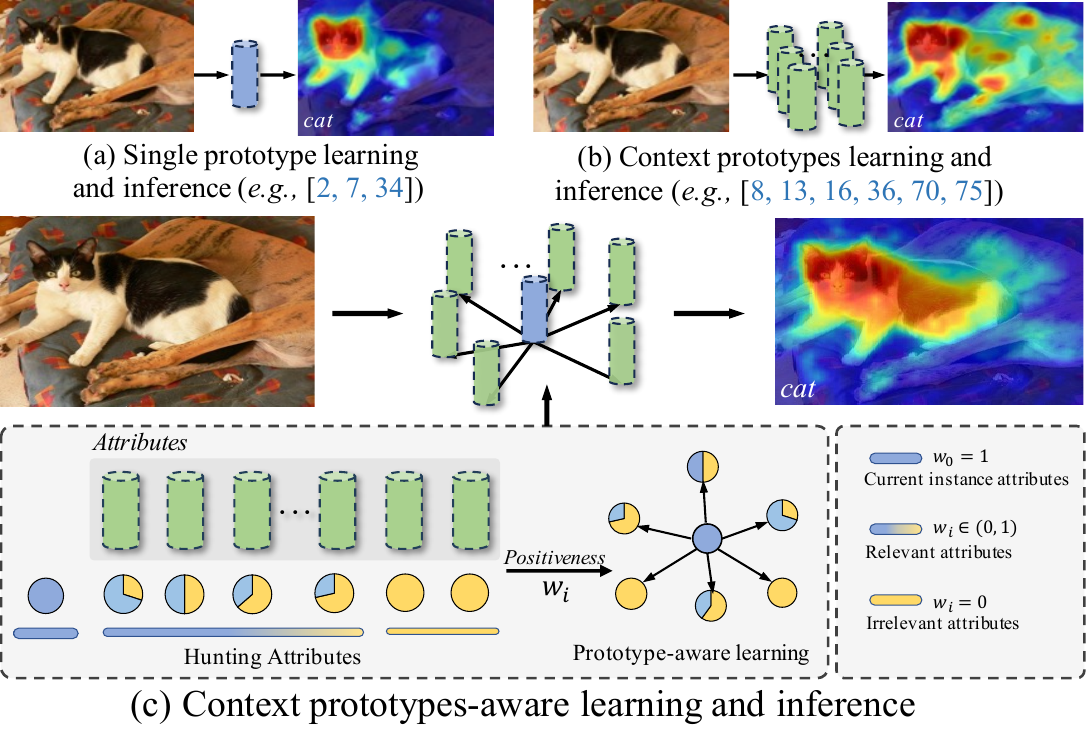}
\end{tabular}
\vspace{-0.4cm}
\caption{
The main idea promoted throughout the paper is that semantic context prototype-aware underpins localization of individual objects in WSSS. Our CPAL performs adaptive perception of diverse attributes (\textit{e.g.,} \texttt{cat}) with attribute hunting (c) rather than from single prototype (a) and plain context prototypes (b). This attribute-specific adaptation not only mitigates the risk of errors where (b) mistakenly identifies similar categories (\textit{e.g.,} \texttt{dog}) but also ensures accurate activation of the complete object region.
} 
\vspace{-0.3cm}
\label{intro}
\end{figure}

The key in WSSS is generating CAM with better coverage on the complete object. Recent studies~\cite{chang2020weakly,sun2020mining,zhang2020inter,wang2023hunting} primarily aim to optimize model segmentation accuracy and stability by integrating contextual knowledge. Inspired by the progress of representation learning~\cite{fan2020learning,wu2021embedded}, some studies~\cite{li2021group,su2021context,zhang2020causal,zhang2022multi} introduce semantic context and instance knowledge for global-scale context modeling to more accurately parse the semantic features of instances. But they ignore the challenge of large intra-class variation i.e., regions belonging to the same class may exhibit a very different appearance even in the same picture. The bias between contextual knowledge (global in-class features) and instance-specific knowledge (unique features) makes the labels propagation hard from image-level to pixel-level. In this work, we argue that alleviating the knowledge bias between instances and contexts can capture more accurate and complete regions. Further, we incorporate extra supervised signals to expedite the alleviation of knowledge biases. 

Class prototype representation, by diminishing the bias, has shown its potential reveal feature patterns in few-shot learning algorithms such as BDCSPN \cite{liu2020prototype}. Prototype learning theory \cite{zhou2022rethinking,wang2019panet} states that prototypes can represent local features, global features, or specific attributes of an object. Based on intra-class variation in object features, an instance prototype \cite{chen2022self} can dynamically characterize
the discriminative features of the specific image. As shown in Fig.~\ref{intro} (a), only few pixels are activated in warm colors, indicating that a large amount of pixels representing the object are mistakenly classified as background. Furthermore, prototypes that integrate contextual knowledge \cite{zhou2022regional} have the ability to capture more specific and accurate category semantic patterns. They enable a more complete capture of the object area compared to a single instance prototype (Fig.~\ref{intro} (b)). Though the introduction of contextual knowledge enhances the ability of prototypes to process semantic information, knowledge bias between instances and contexts result in prototypes erroneously activating similar or highly co-occurring categories (\textit{e.g.,} \texttt{cat} and \texttt{dog} in Fig.~\ref{intro} (b)).

In this work, we propose a learning strategy named Context Prototype-Aware Learning (CPAL) to mine effective feature attributes from the cluster structure of contexts (Fig.~\ref{intro} (c)). Specifically, we explore other instances related to the specific image to construct contextual prototypes as candidate neighbors. Then, in-class attribute hunting is conducted in the candidate neighbor set, locating the current instance prototype as an anchor. Meanwhile, we design a pairwise positiveness score indicative of the correlation between attributes, aiming to identify contextual prototypes (\textit{i.e.,} soft neighbors) highly related to the current attribute. After applying respective positiveness score, the contributions of these prototypes to the anchored instance were dynamically adjusted, thus explicitly mitigating biases associated with intra-class diversity and instance attributes.

The core of our method is prototype awareness. We softly measure the distance between the instance prototype and the contextual prototype to perceive the instance attributes. For robust estimation, categorical support banks are proposed to overcome the limitations on mini-batch, so intra-class feature diversity can be observed in a feature-to-bank manner where class distribution can be globally approximated. However, due to the limited quantity of instance features, there is a bias relative to the feature distribution of the context, affecting the precise awareness of instance. Therefore, we propose feature distribution alignment by introducing a shifting term $\delta$ to the sparse instance features, pushing them towards the dense feature distribution of the categorical support bank. 

In the PASCAL VOC 2012 \cite{everingham2010pascal} and MS COCO 2014 \cite{lin2014microsoft} datasets, we evaluate our method in various WSSS settings, where our approach achieves state-of-the-art performance. The contributions are summarized as follows:
\begin{itemize}
\item We propose a context prototype-aware learning (CPAL) strategy that generates more accurate and complete localization maps by alleviating the knowledge bias between instances and contexts.
\item We propose a feature alignment module combined with dynamic support banks to accurately perceive the attribute of object instances.
\item We propose a unified learning framework consisting of self-supervised learning and context prototype-aware learning, in which two schemes complement each other. Experiments show that our method brings significant improvement and achieves state-of-the-art performances.
% \item Our method is extensively evaluated on two popular WSSS benchmarks by inserting it into multiple WSSS methods.
\end{itemize}

%%%%%%%%% Related Work
\section{Related Work}
\label{sec:Related_Work}
% it still suffers from incomplete and inaccurate activation. A typical drawback of CAM is that it suffers from incomplete and inaccurate activation.

\noindent \textbf{Weakly Supervised Semantic Segmentation} using image-level labels typically generates CAM as a seed for generating pixel-level pseudo labels. A typical drawback of CAM is its incomplete and inaccurate activation. To address this drawback, recent work has proposed various training schemes, such as adversarial erasing \cite{kweon2021unlocking,yoon2022adversarial,sun2021ecs,kweon2023weakly}, region growing \cite{huang2018weakly,wei2018revisiting}, exploring boundary constraints \cite{rong2023boundary,chen2020weakly,lee2021railroad}. The single-image learning and inference model \cite{araslanov2020single,lee2021railroad} focuses on a deeper understanding of the features within an individual image to generate more complete CAM. SIPE \cite{chen2022self} extract customized prototypes multi-scale features to extend coarse object localization maps to obtain the complete extent of object regions.

While past efforts only considered each image individually, recent work focuses on obtaining rich semantic context between different images in the dataset. Recent works \cite{sun2020mining,fan2020cian} address cross-image semantic mining by focusing on capturing pairwise relationships between images. And \cite{li2021group,zhang2022multi,du2022weakly} further perform high-order semantic mining of more complex relationships within a set of images. At the same time, in order to strengthen the representation relationship of the feature space (explore object patterns on the entire data set), RCA \cite{zhou2022regional} introduces a memory bank to store high-quality category features and perform context modeling. CPSPAN~\cite{jin2023deep} proposed to align the feature representation of paired instances under different views, and this alignment was also introduced in the data distribution under different contexts \cite{zhao2023dual}. Unlike previous work on contextual knowledge application, our method can adaptively perceive the semantic attributes and intra-class variations, resulting in more complete CAM activation regions.
\\ \hspace*{\fill} \\
\noindent \textbf{Prototype-based Learning} has been well studied in few-shot \cite{snell2017prototypical,snell2017prototypical}, zero-shot \cite{he2019dynamic} and unsupervised learning \cite{xu2020attribute}. It is worth noting that many segmentation models can be regarded as prototype-based learning networks \cite{wang2019panet, liu2020part, xu2022semi, zhou2022rethinking, ge2023soft}, revealing the possibility of application in image segmentation. \cite{du2022weakly} proposed a prototype-based metric learning method that enforces feature-level consistency in interviews and intra-view regularization. LPCAM~\cite{chen2023extracting} uses prototype learning to also extract rich features of objects. In our work, we learn effective feature attributes within the clustering structure of the context to model diverse object features at a fine-grained level. 

% \clearpage

%%%%%%%%% Method
\section{Methodology}
\label{sec:Method}

\begin{figure*}[t]
  \centering
  \begin{tabular}{cc}
    \includegraphics[width=0.97\textwidth]{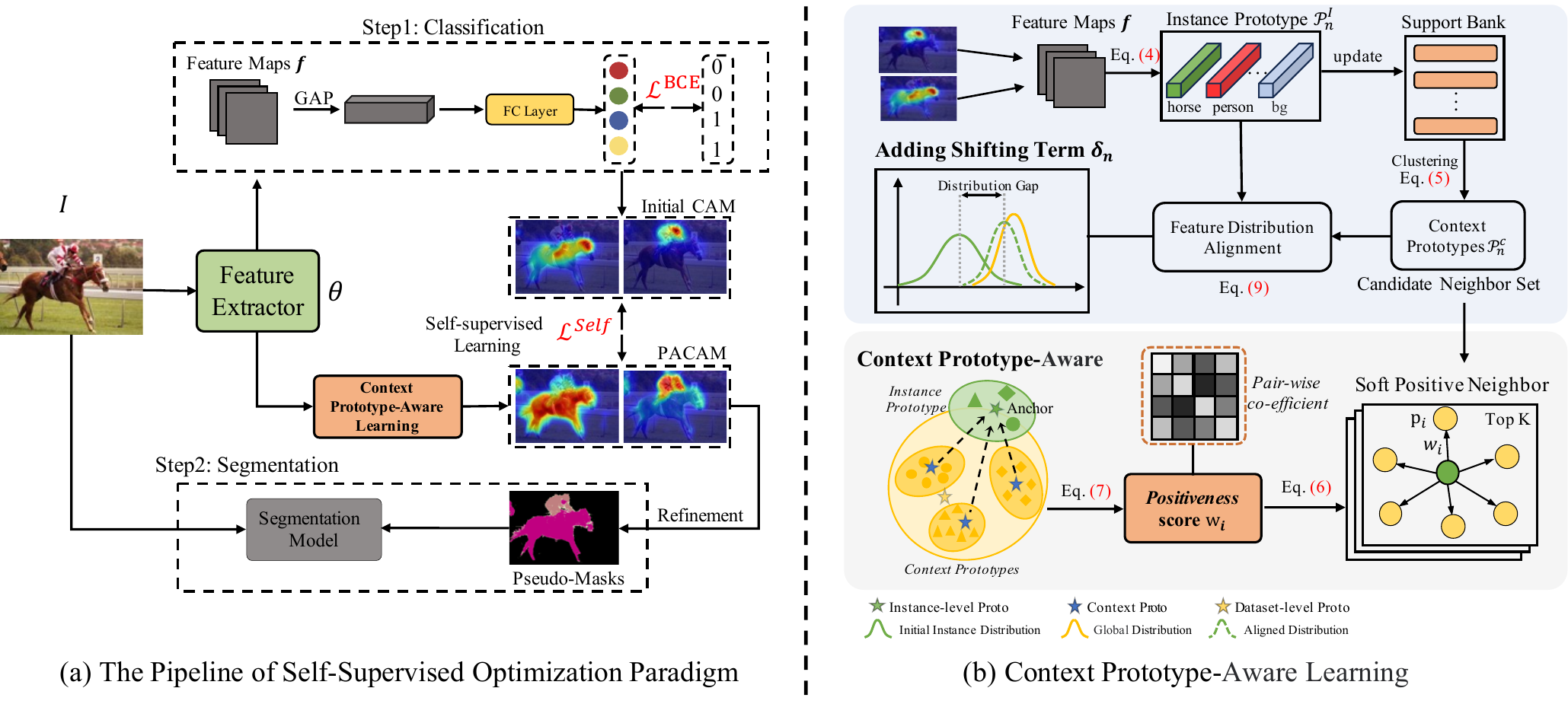} 
  \end{tabular}
  \vspace{-0.2cm}
  \caption{
    Overview of the proposed unified learning framework. 
    (a) shows image label-guided WSSS (\textit{from classification to segmentation}). The upper branch describes the classification network $\theta$ identifying object regions corresponding to each category to minimize $\mathcal{L}^{BCE}$. Introduce a self-supervised learning paradigm using context prototype-aware learning to provide a more complete CAM, supervising the initial CAM and minimizing $ \mathcal{L}^{Self}$. The lower branch refines these CAMs (\textit{e.g.,} DenceCRF \cite{krahenbuhl2011efficient}) to form pseudo-labels for supervising the semantic segmentation network. (b) outlines our strategy based on context prototype-aware learning. In mini-batches, instance prototypes $\mathcal{P}_n^I$ are generated using CAM and extracted features $f$, updating the support bank. Then, the bank is used to construct a context prototype set $\mathcal{P}_n^\text{c}$. Feature distribution alignment is then applied to the current instance features, adding a shift term $\delta_n$ to guide them toward clusters of dense features in the bank. Next, soft neighbors are softly measured for $\mathcal{P}_n^I$ based on $\mathcal{P}_n^\text{c}$, with $\mathcal{P}_n^I$ serving as anchors. Finally, \textit{positiveness} value $w_i$ can be computed between two specific attributes. This mechanism selects $K$ soft positive neighbors $\tilde{\mathcal{P}}_n^{\text {c}}$ to generate PACAM.
      }
      \vspace{-0.3cm}
  \label{method}
\end{figure*}

\quad WSSS first trains the classification network to identify the object region corresponding to each category, then refined to generate pseudo-labels as supervisors of the semantic segmentation network. Fig.~\ref{method} illustrates the overview of the proposed method. The framework is built upon the foundation of a classification network, shown in Fig.~\ref{method} (a) and described in Section~\ref{3.1}. It consists of two supervisory signals: classification loss and self-supervised loss. Our approach encourages consistency between the CAM predicted through prototype-aware learning and the classifier, implicitly motivating the model to learn more discriminative features. We model the instance prototype as an anchor and extract context prototypes from the support bank as the candidate neighbor set, which is described in Section~\ref{3.2}. The core of our method is prototype awareness to capture the intra-class variations, illustrated in Fig.~\ref{method} (b) and detailed in Section~\ref{3.3}. We softly measure the positiveness of each candidate neighbor on the current instance, selectively filter, and adjust their contributions. Meanwhile, feature distribution alignment guides the current instance features toward the cluster center of dense features in the bank. 

\subsection{Self-Supervised Optimization Paradigm}\label{3.1}
\noindent \textbf{Network Optimization.} Our framework is built upon a classification network, utilizing this network $\theta$ to extract effective supervision from image labels, capturing object regions for each category ($i.e.,$ CAMs). We propose context prototype-aware learning to generate more complete prototype-aware CAM (PACAM), providing additional supervisory signals for the initial CAM and forming a self-supervised paradigm. The key element of this paradigm is consistency regularization, implicitly reducing the feature distance between discriminative and missing pixels, encouraging the model to learn more consistent and distinctive features. This simple modification leads to significant improvements. A unified loss function optimizes the model:
\begin{equation}
\label{coefficients}
\mathcal{L}=\lambda_{BCE}\mathcal{L}^{{BCE}}+\lambda_{Self}\mathcal{L}^{ {Self}}
\end{equation}
where $\lambda_{BCE}$ and $\lambda_{Self}$ are coefficients, $\mathcal{L}^{{BCE}}$ is the classification loss, and $\mathcal{L}^{ {Self}}$ is the self-supervised loss. Losses are described in detail in the following sections. 
\\ \hspace*{\fill} \\
\noindent \textbf{Classification Loss and Class Activation Maps.} Each training image $I \in \mathbb{R}^{w \times h \times 3}$ in the dataset $\mathcal{I}$ is associated with only an image-level label vector $\boldsymbol{y}=\{y_n\}^N_{n=1} \in \{0,1\}^N$ for $N$ is pre-specified categories. CAM is proposed to locate the foreground objects by training a classification network. CAM takes a mini-batch image $I$ as the input to extract feature maps $f\in \mathbb{R}^{W \times H \times D}$, with $D$ channels and $H \times W$ spatial size. To bridge the gap between the classification task and the segmentation task,  a classifier weight $\mathbf{w}_n$ and a global average pooling (GAP) layer are employed to produce the logits prediction $\hat{y}_i \in \mathbb{R}^N$. During training, binary cross-entropy loss is used as follows: 
\begin{equation} \mathcal{L}^{BCE}=\frac{1}{N} \sum_{i=1}^N y_i \log \sigma\left(\hat{y}_i\right)+\left(1-y_i\right) \log \left(1-\sigma\left(\hat{y}_i\right)\right), 
\end{equation}
where $\sigma(\cdot)$ is the sigmoid function. To get rough location information about foreground and background. The class activation map ${M}_{\boldsymbol{f}}=\left\{{M}_n\right\}_{n=1}^N$ over $N$ foreground classes can be represented as follows:
\begin{equation} {M}_{n}=\frac{\operatorname{ReLU}\left(\boldsymbol{\mathbf{w}_n^{\top} f}\right)}{\max \left(\operatorname{ReLU}\left(\boldsymbol{\mathbf{w}_n^{\top} f}\right)\right)}, \quad \forall n \in N.
\end{equation}
Considering the importance of background in the segmentation task, we follow \cite{wang2020self} to estimate the background activation map ${M}_{b}=1-\max_{1 \leq n \leq N} M_n$ based on $M_f$. We combine the processed background activation map with the foreground activation map as a whole, \textit{i.e.} ${M} = M_f \cup M_b$, to help model background knowledge.

\subsection{Prototype Modeling}\label{3.2}
\quad Inspired by prototype-based learning, our prototype awareness strategy aims to effectively explore features within the candidate neighbor set. We propose to conduct a prototype search within the context prototype set for each class, locating the current instance prototype as an anchor to enhance comprehension of the instance features. 
\\ \hspace*{\fill} \\
% 在本节中，我们将详细描述实例原型和上下文原型的建模过程，强调它们之间的关系。实例原型更专注于捕获特定图像的判别性特征，而上下文原型通过对整个数据集的聚合提供了更全局、更综合的类别语义信息。实例原型和上下文原型的协同作用使得网络能够更全面、准确地捕捉不同实例图像的语义特征。根据第3.4.1节的密度估计，我们对每个小批量中的分类特征执行密度引导的锚点和键选择。 
% 受密度峰值假设[33]的启发，我们的对比学习策略旨在通过将稀疏区域的特征推向由密集聚集的特征近似的类中心来提高类内聚类的紧凑性。我们建议使用特征密度作为指标来决定应该选择哪些特征作为锚点。我们在每个小批量中对分类特征执行密度引导锚点和键选择。Inspired by density-peak assumption [33], our contrastive learning strategy aims to increase in-class cluster compact- ness by pushing features in sparse regions towards the class center approximated by densely gathered features. We propose to use feature density as an indicator to decide which feature should be selected as anchors. In-batch features can provide fresh and in-object contrast, while the global memory bank shows a more comprehensive and diversified categorical pattern.
% 受原型学习的启发，我们的感知原型策略旨在将候选邻居集中挖掘有效的特征属性。我们提出在每个类内上下文原型集中进行属性搜索，定位当前实例原型作为锚点，从而增强对实例属性的理解。每个原型代表在单个图像中观察到的类别的区域语义。Each prototype represents the regional semantics of the categories observed in $I$
% Each prototype represents the region-aware semantic of the $n$-th category observed in individual images. we convert the projected features into categorical region representations based on $M$.
\noindent\textbf{Modeling Instance Prototype as Anchor.} For each image $I$, feature maps are mapped to the projection space $z=v(f)$ by a projection head $v$ for instance prototyping. Each instance prototype represents the regional semantics of the categories observed in $I$ based on $M$. Specifically, for the $n$-th category that appears in $I$ ($i.e.,$ $y_c=1$), its projected features is summarized to a vector $\mathcal{P}^{I}_n \in \mathbb{R}^D$ by masked average pooling (MAP) \cite{siam2019amp}:
\begin{equation}
\mathcal{P}^{I}_n=\frac{\sum_{x=1, y=1}^{W, H} \textbf{P}_n(x,y) * z(x,y)}{\sum_{x=1, y=1}^{W, H} \textbf{P}(x,y)}, 
\label{tau}
\end{equation}
where $\textbf{P}_n= \mathbbm{1}\left({M}_n>\tau\right) \in \{0,1\}^{W \times H}$ is a binary mask, emphasizing only strongly-activated pixels of class $n$ in its activation map. $\mathbbm{1}(\cdot)$ is an indicator function, and the threshold $\tau$ is a hyper-parameter and denotes the threshold of the reliability score. Here, $\mathcal{P}^{I}_n$ is compact and lightweight, allowing feasible exploration of its relationships with numerous other samples and positioning it as an anchor.
% This process performs class-wise compression on the region feature, achieving N foreground prototypes and one background prototype. 在这里，\mathcal{P}^{I}_n是紧凑和轻量级的，允许可行地探索它与从大量其他样本中之间的关系，并将该实例原型作为锚点。
% \vspace{-0.1cm}
\\ \hspace*{\fill} \\
\noindent\textbf{Modeling Context Prototypes as Candidate Neighbors.} We assume that categorical features within images or batches only provide a limited view of the class. Therefore, we utilize a support bank as a candidate set $\mathcal{C}$, where each element is the context prototype of different categories. When using sample batches for network training, we store their instance prototypes $\mathcal{P}^{I}_n$ in $\mathcal{C}$ and employ a first-in-first-out strategy to update the candidate set. This set maintains a relatively large length for each prototype category to sufficiently provide potential context prototypes. Based on this set, $\mathrm{k}$-means online clustering is applied to refine each category into clustered prototype groups $\mathcal{G}=\left\{G_i\right\}_{i=1}^{N_p}$ to deeply reveal attributes of each category. We perform averaging operations on each clustered prototype group of $\mathcal{G}$ to generate $N_p$ candidate neighbors $\mathbf{p}_i$ as follows:
\begin{equation}  
\mathbf{p}_i=\frac{1}{\left|G_i\right|} \sum_{\mathbf{r}_j \in G_i} \mathbf{r}_j, 
% \vspace{-0.2cm}
\end{equation} 
where $\mathbf{r}_j$ refers to the $j$-th instance prototype belonging to the $i$-th cluster group $G_i$.
$\mathbf{p}_i$ represents the $i$-th context prototype of the candidate neighbor set $\mathcal{P}_n^c=\left\{\mathbf{p}_i\right\}_{i=1}^{N_p}$.
% where $\mathbf{p}_i$ is the average vector of $\mathbf{r}_j$ instances, where each $\mathbf{r}_j$ belongs to the $i$-th cluster group $G_i$.
\subsection{Context Prototype-Aware Learning}\label{3.3}
\quad With the anchor prototypes and the candidate neighbor set from Section~\ref{3.2},  the candidate neighbor set further perceives or supports the anchor feature. Context prototype-aware learning can measure and adjust this support extent.
\\ \hspace*{\fill} \\
\noindent\textbf{Soft Positive Neighbor Identification.} 
\quad Prototype selection is critical in our proposed approach since it largely determines the quality of supervision. Instance prototypes can specifically represent the categorical attributes of the current image, while context prototypes show more comprehensive and diverse category patterns. Our awareness strategy employs positiveness scores $w_i$ to measure the relevance of candidate neighbors in the category to the current instance attributes. We propose selecting the top-$K$ neighbors adjusted by positiveness scores, located in close proximity to the anchor. The soft positive neighbor can be formulated as:
\begin{equation}
\tilde{\mathcal{P}}_n^{\text {c}}=\left\{w_i \mathbf{p}_{\mathbf{i}}: i \in \underset{i \in N_p}{ \arg \max }\left(d\left(w_i \mathbf{p}_{\mathbf{i}}, \mathcal{P}_n^I\right), \text { top } =K\right)\right\}
\label{value_K}
\end{equation}
where $d()$ denotes the cosine similarity as the measured metric, and $\tilde{\mathcal{P}}^{c}_n$ represents the top-$K$ context-aware prototypes tailored to the current instance. 
% \\ \hspace*{\fill} \\

% 这句可要可不要This approach ensures that the model focuses more on a small subset of prototypes that are similar to the current instance, enhancing the accuracy of the model in recognizing and expressing specific category attributes. This mechanism proves effective in handling datasets with complex and similar categories, improving the model's performance and generalization capabilities.
\noindent \textbf{Positiveness Predictions.} We have designed pair-wise positiveness scores to softly measure (in a non-binary form) the relevance between the instance prototype and the candidate neighbors in the same category. For the prototype pair ($\mathbf{p}_i$ , $\mathcal{P}^I_{n}$), the positiveness score $w_{i}$ can be calculated as:
\begin{equation} 
w_i=\frac{1}{\gamma_i} \texttt{softmax}\left[l_1\left(\mathbf {\mathcal{P}}^{I}_n\right) \times l_2\left(\mathbf {p}_i\right)^{\top}\right], \quad {\mathbf{p}}_{i} \in {\mathcal{P}}^{c}_n, 
\label{eq7}
\end{equation}
where $l_1(\cdot)$ and $l_2(\cdot)$ are parameter-free identity mapping layers in feature transformation. $\gamma_i$ is a scaling factor to adjust the positiveness score $w_i$. Various structures for the score $w_{i}$ have been explored in Section \ref{Ablation}. 
% \vspace{-0.25cm}
\\ \hspace*{\fill} \\
\noindent\textbf{Claim 1.} \textit{Assume we train a model $\theta$ using the proposed optimization method, $\mathcal{P}_n^I$ and $\tilde{\mathcal{P}}_n^c$ are n-th class current instance prototype and context prototypes, respectively.  The optimal value of similarity measure $s_i^*$ can be expressed as $\frac{w_{i}}{\sum_{k=1}^{K} w_{k}}$, where $w_{i}$ is the corresponding positiveness score for the prototype pair  ($\mathcal{P}_n^I, \quad {\mathbf{p}}_{i} \in {\tilde{\mathcal{P}}}_n^{c}$) in Eq.~\ref{eq7}.} 
% \vspace{-0.25cm}
\\ \hspace*{\fill} \\
\noindent The proof can be found in Appendix A. Claim 1 indicates that we optimize the model to maximize the similarity between the context prototype and the current instance of the same category in direct proportion to the corresponding positiveness score. We effectively transfer knowledge from the self-supervised branch to the model, as well as the generalization performance of the model. 
% \vspace{-0.25cm}
\\ \hspace*{\fill} \\
\noindent\textbf{Feature Distribution Alignment.}
The sparse features~\cite{hoefler2021sparsity} and intra-class diversity pose challenges to accurately representing consistent category-specific features, impeding category distinction. Thus, we posit bias between instance and intra-class features. To tackle this, we guide features to align their category-specific densely gathered features to enhance intra-class feature compactness. Considering that mini-batch normalization \cite{ioffe2015batch} or instance normalization \cite{ulyanov2016instance} follows the trend of batch learning, the mini-batch features are aligned by introducing shift terms $\delta_n$ to push them towards the cluster centers. The derivation is as follows.

We define the Optimal Cosine Similarity Evaluation Metric (OCSEM) to assess the cosine similarity between the current sample and others, aiming to boost model accuracy by maximizing this metric. The optimization objective is defined as:
\begin{equation}
\begin{split}
\text{OCSEM} = \frac{1}{{N_p}{Q_n}} \sum^{{N_p}}_{i=1} \sum^{{Q_n}}_{q=1} & \cos({\mathbf {p}}_{i},\mathcal P^I_{n,q}) > \\ &  \max_{h \neq i}\{\cos({\mathbf {p}}_{h},\mathcal P^I_{n,q})\}, 
\end{split}
\end{equation}
where ${\mathbf {p}}_{i}$ is the context prototype in the candidate neighbors set $\mathcal{P}_n^c=\left\{\mathbf{p}_i\right\}_{i=1}^{N_p}$ for the $n$-th class, and $\mathcal P^I_{n,q}$ is its corresponding instance prototype in the set $\mathcal{P}_n^b=\left\{\mathcal P^I_{n,q}\right\}_{q=1}^{Q_n}$ in the mini-batch. $Q_n$ denotes the number of prototypes for the $n$-th class in the mini-batch. We assume the bias can be diminished by adding a shifting term $\delta_n$ to the instance feature. The term $\delta_n$ should follow the objective:
\begin{equation}
\underset{\delta_n}{\arg \max } \frac{1}{{N_p}{Q_n}} \sum_{i=1}^{N_p} \sum_{q=1}^{Q_n} \cos \left({\mathbf {p}}_{i}, \mathcal P^I_{n,q}+\delta_n\right).
\label{9}
\end{equation}
We assume that each prototype features $\mathcal P^I_{n,q}$ can be represented as ${\mathbf {p}}_{i} + \epsilon_{i,q}$. Eq.~\ref{9} can be further formalized as:
\begin{equation}
\underset{\delta_n}{\arg \max } \frac{1}{{N_p}{Q_n}} \sum_{i=1}^{N_p} \sum_{q=1}^{Q_n} \cos \left({\mathbf {p}}_{i}, {\mathbf {p}}_{i}+\delta_n+\epsilon_{i, q}\right).
\end{equation}
To maximize the cosine similarity, we should minimize the following objective:
\begin{equation}
\min \frac{1}{{N_p}{Q_n}} \sum^{{N_p}}_{i=1} \sum^{{Q_n}}_{q=1} (\epsilon_{i,q}+\delta_n).
\end{equation}
The term $\delta_n$ is thus computed:
\begin{equation}
\delta_n=-\mathbb{E}\left[\epsilon_{i,q}\right]=\frac{1}{{N_p}{Q_n}} \sum_{i=1}^{N_p} \sum_{q=1}^{Q_n} \left({\mathbf {p}}_{i}-\mathcal{P}_{n, q}^I\right).
\label{shift}
\end{equation}

\subsection{Prototype-Aware CAM and Self-Supervise Loss}

\noindent \textbf{Prototype-Aware CAM.} 
With the clear meaning of the prototypes, the predicted CAM procedure can be intuitively understood as retrieving the most similar prototypes. For each prototype  $\tilde{\mathcal{P}}^{c}_n$ in Eq.~\ref{value_K}, we compute the cosine similarity between features at each position and the corresponding category prototype. These similarity maps are then aggregated as follows:
\begin{equation}  {\tilde{M}}_n(j) = \ ReLU \left(\frac{1}{K} 
\sum_{{\mathcal{\mathbf p}}_i \in {\tilde{\mathcal{P}}}^{c}_n}
\frac{{{f}}{(j)} \cdot {\mathcal{\mathbf p}}_i}{\left\|{{f}}(j)\right\| \cdot\left\|{\mathcal{\mathbf p}}_i\right\|}\right),
\end{equation}
where $\|\cdot\|$ denotes the L2-norm of a vector. $\tilde{M}_n(j)$ represents the PACAM for the $n$-th class at pixel $j$. 

\noindent \textbf{Self-Supervise Loss.} To further leverage contextual knowledge, we introduce a self-supervised learning paradigm that encourages consistency between outputs from prototype-aware predictions and a supervised classifier. This promotes the model to recognize more discriminative features and injects prototype-aware knowledge into the feature representation, fostering collaborative optimization throughout training cycles. The L1 normalization of two CAMs defines the consistency regularization:
\begin{equation} 
\mathcal{L}^{self}=\frac{1}{N+1}\|{M}- {\tilde{M}}\|_1,
\label{self}
\end{equation}
where $M$ and $\tilde{M}$ represent the original CAM and PACAM, respectively.
% In this work, we argue that alleviating the knowledge bias between instances and contexts can capture more accurate and complete regions. Further, we incorporate extra supervised signals to expedite the alleviation of knowledge biases. 

\begin{table}[t]
\centering
\scriptsize
\caption{Ablation study on main components of the proposed framework. The mIoU values are evaluated on the PASCAL VOC 2012 \cite{everingham2010pascal} \texttt{train} set. $\mathcal{L}^{BCE}$: Baseline classification network. Vanilla: Plain context learning, where context prototypes are clustered from the support bank. Proto-aware: prototype-aware learning involving the top-$K$ candidate neighbor set and positiveness prediction in Eq.~\ref{value_K}. Align: Feature alignment, adding shift term within the mini-batch feature in Eq.~\ref{shift}. $\mathcal{L}^{Self}$: Self-supervised loss used as an additional supervised signal in Eq.~\ref{self}.}
\vspace{-0.25cm}
\begin{tabular}{c|c|c|c|c|c|c}
\toprule[1pt] & $\mathcal{L}^{{BCE}}$ & Vanilla & Proto-Aware & $\text {Align}$ & $\mathcal{L}^{{Self}}$  & mIoU \\  \midrule\midrule
\text { I } & \CheckmarkBold & & & & & 50.1  \\
\text { II } & \CheckmarkBold & \CheckmarkBold & & & & 51.2 \\
\text { III } & \CheckmarkBold & \CheckmarkBold & \CheckmarkBold & & & 54.5 \\
\text { IV } & \CheckmarkBold & \CheckmarkBold & \CheckmarkBold & \CheckmarkBold & & 56.8 \\
\text { V } & \CheckmarkBold &  \CheckmarkBold & \CheckmarkBold & \CheckmarkBold & \CheckmarkBold & 62.5 \\
\toprule[1pt]
\end{tabular}\label{abl}
\vspace{-0.4cm}
\end{table}

\section{Experiments}
\label{sec:Experiments}
\subsection{Datasets and Implementation Details}
\textbf{Dataset and Evaluation Metric.} Experiments are conducted on two benchmarks: PASCAL VOC 2012 \cite{everingham2010pascal} with 21 classes and MS COCO 2014 \cite{lin2014microsoft} with 81 classes. For PASCAL VOC 2012, following \cite{wang2020self,lee2021anti,chen2022self,li2022expansion}, we use the augmented SBD \cite{hariharan2011semantic} with 10,582 annotated images. We evaluate CPAL in terms of i) quality of generated pseudo segmentation labels on VOC 2012 \texttt{train}, and ii) semantic segmentation on VOC 2012 \texttt{val/test} and COCO 2014 \texttt{val}. Mean intersection over union (mIoU) \cite{long2015fully} is used as the metric in both cases. The scores on the VOC 2012 \texttt{test} are obtained from the official evaluation server.

\noindent\textbf{Implementation Details.} In our experiments, the ImageNet~\cite{deng2009imagenet} pre-trained ResNet50~\cite{he2016deep} is adopted as the backbone with an output stride of 16, where a classifier replaces the fully connected layer with output channels of 20. The augmentation strategy is the same as \cite{chen2022self,ahn2019weakly,chen2023extracting}, including random flipping, scaling, and crop. The model is trained with a batch size 16 on 8 Nvidia 4090 GPUs. SGD optimizer is adopted to train our model for 5 epochs, with a momentum of 0.9 and a weight decay of 1e-4. The learning rates for the backbone and the newly added layers are set as 0.1 and 1, respectively. We use a poly learning scheduler decayed with a power of 0.9 for the learning rate.

The loss coefficients $\lambda_{BCE}$ and $\lambda_{Self}$ are both set as 1 in Eq.~\ref{coefficients}. For VOC 2012, the threshold $\tau$ in Eq.~\ref{tau} is set to 0.1. Support bank size for each class to store region embeddings, with the size set to 1000 to avoid significant support consumption. The $k$-means prototype clustering in Section~\ref{3.2} is performed only once at the beginning of each epoch, and the per-class prototype number $N_p$ is set to 50, and the top-$K$ candidate neighbors is set to 20 in Eq.~\ref{value_K}. For the segmentation network, we experimented with DeepLab-v2~\cite{chen2017deeplab} with the ResNet101 and ResNet38 backbone. \textit{More details (including COCO) are in the appendix.}

% \caption{Ablation study on main components of the proposed framework. The mIoU values are evaluated on the PASCAL VOC 2012 \cite{everingham2010pascal} \texttt{train} set. $\mathcal{L}^{BCE}$: Baseline classification network. Vanilla: Plain context learning, where context prototypes are clustered from the support bank. Proto-aware: the top-K candidate neighbor set and positiveness prediction introduce context prototype learning in Eq.~\ref{value_K}. Align: Feature alignment, adding shift term within the mini-batch feature in Eq.~\ref{shift}. $\mathcal{L}^{Self}$: Self-supervised loss used as an additional supervised signal in Eq.~\ref{self}.}

% , which thus alleviates the large intra-class variations bias and encourages the prototypes to capture more discriminative features.
\begin{figure}[t]
\centering
\begin{tabular}{cc}
\includegraphics[width=0.46\textwidth]{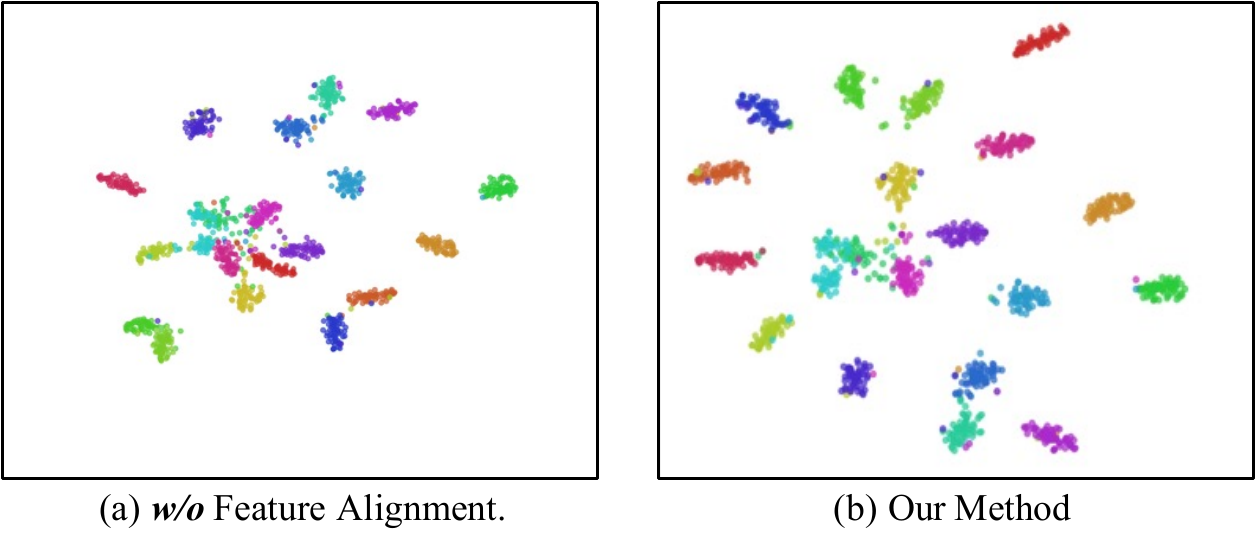}
\end{tabular}
\vspace{-0.4cm}
\caption{Feature embedding visualizations of (a) our method without feature distribution alignment, and (b) our method on the Pascal VOC 2012 \texttt{val} images using t-SNE~\cite{van2008visualizing}. Feature distribution alignment improves the compactness of intra-class features.} 
\vspace{-0.2cm}
\label{tsnet}
\end{figure}

\begin{figure}[t]
\centering
\begin{tabular}{cc}
\includegraphics[width=0.46\textwidth]{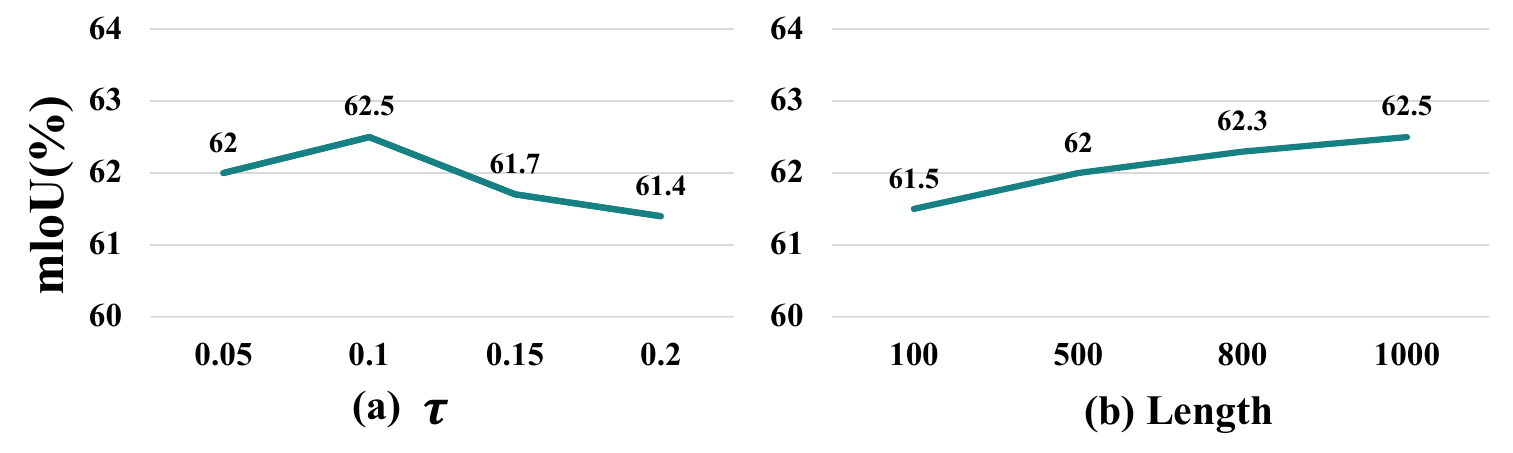}
\end{tabular}
\vspace{-0.1cm}
\caption{Sensitivity analysis on PASCAL VOC 2012 \texttt{train} set, in terms of (a) the threshold $\tau$ used to generate 0-1 seed masks from heatmaps. (b) the length of the support set. The results show that CPAL is not sensitive to them.} 
\vspace{-0.4cm}
\label{hyperparameter}
\end{figure}

% The performance tends to be saturated when the length of prototypes reaches 1000 for each class. The performance is relatively stable with variations of parameters. The support set is constructed to support the current instance and is used to model the context prototypes.

\begin{figure*}[t]
\centering
\begin{tabular}{cc}
\includegraphics[width=0.99\textwidth]{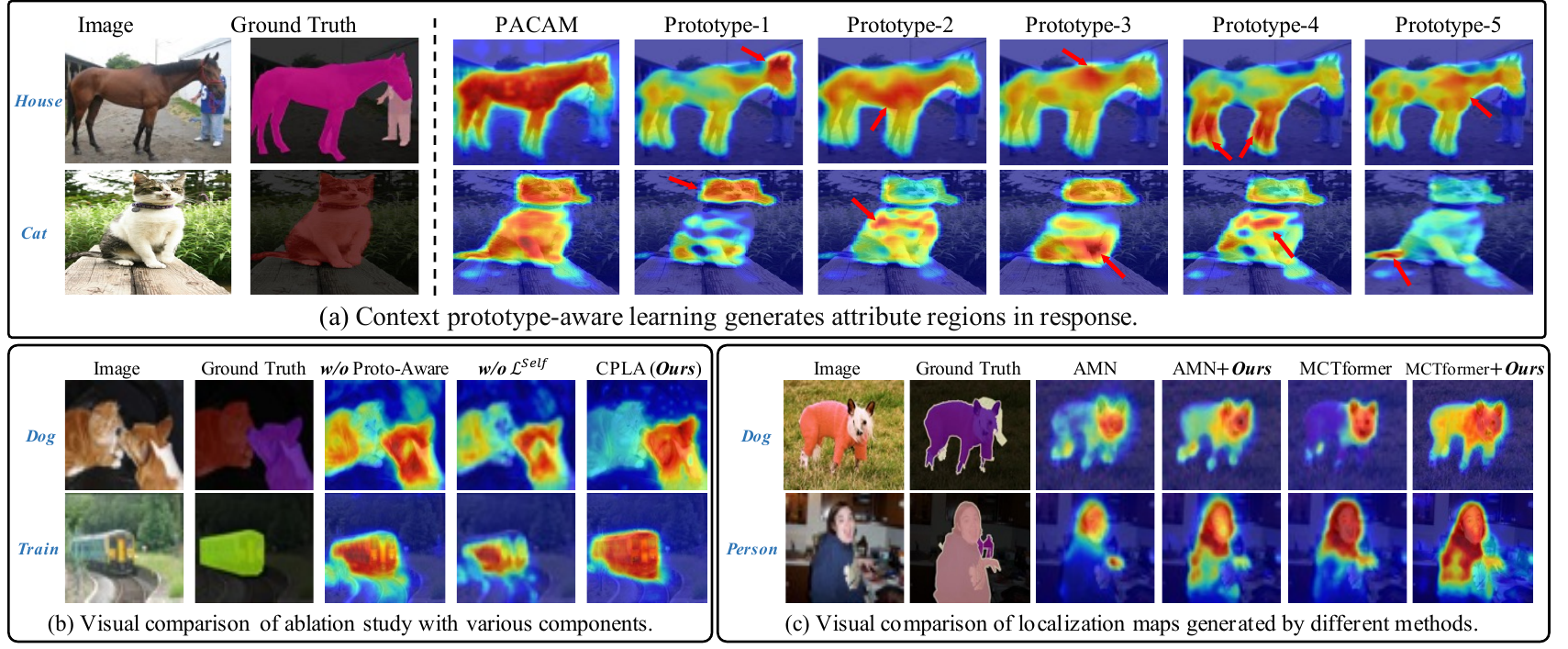}
\end{tabular}
\vspace{-0.3cm}
\caption{Qualitative visualization on the PASCAL VOC \texttt{train} set. (a) PACAM is obtained using various soft positive prototypes to enhance the comprehension of our model. (b) Visual comparison of ablation study two main components: our model without prototype-aware learning (top-$K$ candidate neighbor set and positiveness prediction) or self-supervised loss. (c) The impact of our method as a plug-in to AMN~\cite{lee2022threshold} and MCTformer~\cite{xu2022multi} significantly improves the object localization ability of networks.} 
\vspace{-0.3cm}
\label{fig31}
\end{figure*} 

% It showcases the ability of our model to activate accurate and complete localization maps, addressing co-occurrence ($e.g.,$ \texttt{train} and \texttt{railroad}) and similar categories ($e.g.,$ \texttt{dog} and \texttt{cat}) issues. 

\subsection{Ablation study}\label{Ablation}
To study the contributions of each component of our method, we conducted ablation studies on the VOC 2012 dataset. All experiments use Resnet-50 as the backbone.
% \textbf{Cluster Performance Comparison}: One key observation is that our Context Prototype-Aware Learning (CPAL) has learned better representations in the feature space. We performed ablation studies on PASCAL VOC 2012 to demonstrate its effectiveness both visually and quantitatively. We extracted clusters from the validation set of PASCAL using models trained on it. Feature categories are assigned based on the model's predictions. Figure 1 compares the feature space learned by the original CAM baseline and CPAL. Evidently, our method generates more separable clusters. In Table 4, we evaluated various aspects of cluster performance using multiple metrics such as Silhouette, Calinski-Harbasz, and Davies-Boulding scores. The results indicate that CPAL significantly improves the baseline's performance, creating more compact clusters with higher inter-cluster separability.
% \vspace{-0.25cm}
% \\ \hspace*{\fill} \\
\noindent\textbf{Effectiveness of each component.}
In Table~\ref{abl}, we conduct ablation studies to demonstrate the effectiveness of our approach. We utilize a model trained with only classification supervision training (Experiment I) as the baseline. Then, a plain context prototype learning strategy is introduced in Experiment II and only brings limited gains in mIoU on the \texttt{train} set. Experiment III demonstrates that introducing context prototype-aware learning (top-$K$ candidate neighbor set and positiveness prediction) to generate PACAM significantly boosts performance by +3.3\%. In Experiment IV, when introducing the feature alignment module, performance further increased by +2.3\%. In Experiment V, the performance is further improved by +5.7\% when introduced for self-supervised training as complement supervision, indicating its importance in our framework. The consistency loss compels the model to concentrate on fine-grained semantic details, enhancing its perception of the intrinsic structure and semantic features. \\
% \vspace{-0.25cm}
% \\ \hspace*{\fill} \\
\noindent \textbf{Effectiveness of candidate neighbors and positiveness.} We analyze the importance of candidate neighbors and positiveness, as shown in Table~\ref{neighbor}. Removing positiveness and utilizing all neighbors for prediction, Miou accuracy decrease in CAM from 62.5\% to 60.3\%. It indicates that positiveness is not merely a simple embellishment but rather provides an effective mechanism for the model. It enables the model to adaptively and selectively focus on neighbors that contribute most significantly to the task during the learning process while disregarding neighbors that are uninformative for predictions. In the third block of Table~\ref{neighbor}, we also conduct experiments to analyze the influence of the number of neighbors. On one hand, having a sufficient number of neighbors enhances the diversity of features. On the other hand, including less-correlated prototypes may introduce too much noise during the training process and diminish the ability of the model to perceive discriminative features. The proposed soft measure introduces a pair-wise positiveness to adjust the contribution of different prototypes to the anchor instance in Eq.~\ref{coefficients}. We apply various similarity metrics to calculate the positiveness score. As illustrated in Table~\ref{function}, four options were explored: Manhattan distance ($L_1$), Euclidean distance ($L_2$), Cosine similarity, and Dot product. The Dot product demonstrates significantly superior performance compared to the other strategies and is used as our method to measure positiveness.

% \begin{table}[t]
% \centering
% \footnotesize
% \caption{Analysis of the positiveness and number of candidate neighbors $K$. The mIoU values are evaluated on the PASCAL VOC 2012 \texttt{train} set.}
% \vspace{-0.25cm}
% \newcolumntype{Y}{>{\raggedleft\arraybackslash}X}
% \begin{tabularx}{0.4\textwidth}{YXXX}
% \toprule[1pt] Neighbor & Positiveness & $K$ & mIou(\%) \\
% \midrule \midrule \CheckmarkBold & \CheckmarkBold & 20 & 62.5 \\
% \midrule \XSolidBrush & \XSolidBrush & - & 59.2 \\
% \CheckmarkBold & \XSolidBrush & 20 & 60.3  \\
% \midrule \CheckmarkBold & \CheckmarkBold & 10 & 61.3 \\
% \CheckmarkBold & \CheckmarkBold & 20 & 62.5 \\
% \CheckmarkBold & \CheckmarkBold & 50 & 60.1 \\
% \toprule[1pt]
% \end{tabularx}
% \vspace{-0.3cm}
% % \caption{This is an example table}
% \end{table}
\begin{table}[t]
\centering
\footnotesize
\caption{Analysis of the positiveness and number of candidate neighbors $K$. The mIoU values are evaluated on the PASCAL VOC 2012 \texttt{train} set.}
\vspace{-0.25cm}
\begin{tabular}{cccc}
\toprule[1pt] Neighbor & Positiveness & $K$ & mIou(\%) \\
\midrule \midrule \CheckmarkBold & \CheckmarkBold & 20 & 62.5 \\
\midrule \XSolidBrush & \XSolidBrush & - & 59.2 \\
\CheckmarkBold & \XSolidBrush & 20 & 60.3  \\
\midrule \CheckmarkBold & \CheckmarkBold & 10 & 61.3 \\
\CheckmarkBold & \CheckmarkBold & 20 & 62.5 \\
\CheckmarkBold & \CheckmarkBold & 50 & 60.1 \\
\toprule[1pt]
\label{neighbor}
\vspace{-0.3cm}
\end{tabular}
\vspace{-0.6cm}
\end{table}
\begin{table}[t]
\centering
\caption{Quantitative comparison of different distance measure strategies in $positiveness$ on PASCAL VOC \texttt{train} set. The best results are shown in bold.}
\vspace{-0.25cm}
\begin{tabular}{c|cccc}
\toprule[1pt] \text { Function } & $L_1$ & $L_2$ & Cosine & Dot \\ \midrule \midrule 
mIou (\%) & 59.6 & 58.7 &  61.9 &  \textbf{62.5}\\
\toprule[1pt]
\end{tabular}
\vspace{-0.8cm}
\label{function}
\end{table}
% \\ \hspace*{\fill} \\
\noindent\textbf{Effectiveness of feature alignment.} In Table \ref{abl}, we present the performance improvement results achieved by diminishing distribution bias. Additionally, we conducted a visual comparison using t-SNE~\cite{van2008visualizing} in Fig.~\ref{tsnet}. The results indicate that after aligning the feature distributions, the model can generate more compact clusters with higher inter-cluster separability. Adjusting the dynamic shift variable helps alleviate differences between instance features of the same class, making instances belonging to the same class more similar. This, in turn, facilitates the model in distinguishing instances from different categories more accurately. \\
% \vspace{-0.35cm}
% \\ \hspace*{\fill} \\ 
\noindent\textbf{Analysis of Hyper-parameters.} We conduct a hyperparameter sensitivity analysis, varying values such as (a) the threshold $\tau$ for generating the 0-1 seed mask. Fig.~\ref{hyperparameter} (a) indicates that the optimal $\tau$ value is 0.1. Additionally, we examine (b) the length of the support set, finding that a larger set enhances model performance. Fig.~\ref{hyperparameter} (b) shows that the encoder trained with the largest set achieves the highest accuracy of 62.5\%, suggesting that increasing capacity enables the model to find more correlated neighbors for support.  \\
% \vspace{-0.25cm}
% \\ \hspace*{\fill} \\
\noindent \textbf{Qualitative Analysis:} We visualize the response regions and prediction outcomes of prototype awareness in Fig.~\ref{fig31} (a). It clearly demonstrates that prototypes are associated with specific instance attributes. Specifically, For example, given images (\textit{e.g.,} \texttt{horse} and \texttt{cat}), each prototype corresponds to different parts of the instance, enabling better modeling of intra-class variations in semantic objects. In Fig.~\ref{fig31} (b), we conduct visualizations of ablation studies on different components of our method. When removing prototype awareness (positiveness and top-$K$ neighbors), the model erroneously activates regions that strongly co-occur (\textit{e.g.,} \texttt{train} and \texttt{railroad}) or exhibit similar appearances (\textit{e.g.,} \texttt{cat} and \texttt{dog}), indicating a lack of accurate learning and discriminative capabilities for instance-specific features. Without self-supervised loss $\mathcal{L}^{Self}$, CAM shows under-activation, indicating insufficient learning of category features. These findings suggest that our method, with the introduction of these components, can more accurately perceive and distinguish various category attributes.
% , ultimately enhancing model performance.

% These findings suggest that our method, with the introduction of these components, can more accurately perceive and distinguish various category attributes, ultimately enhancing model performance. 
% Each prototype corresponds to different parts of the instance, better modeling of intra-class variations of semantic objects.

% Figure~\ref{fig31} presents the qualitative results of the model on the PASCAL VOC 2012 \texttt{train} set. To gain a better understanding of our prototype-aware learning framework, we visualize the response regions of our part prototypes and the prediction results in (a). In (b), we illustrate the effects of our approach by visualizing the without of prototype awareness (positiveness and neighbors) and the without of consistency loss, demonstrating a significant improvement in the quality of localization maps and addressing issues of co-occurrence and similar categories.

\begin{table}[t]
\centering
\footnotesize
\caption{Comparisons between our method and the other WSSS methods. We evaluate mIoU (\%) on the PASCAL VOC 2012 \texttt{train} set at levels: CAM, w/ CRF, and pseudo Mask.}
\vspace{-0.25cm}
\begin{tabular}{lccc}
\toprule[1pt]
Method & Seed & w/ CRF  & Mask \\ 
\midrule  
% $\textnormal{IRN}~{\resizebox{0.9cm}{!}{\textnormal{\textcolor{gray}{[CVPR19]}}}}$ ~\cite{ahn2019weakly} & 48.8 & 53.7 & 66.3 \\
$\textnormal{SEAM}~{\resizebox{0.9cm}{!}{\textnormal{\textcolor{gray}{[CVPR20]}}}}$ 
~\cite{wang2020self} & 55.4 & 56.8 & 63.6\\
$\textnormal{AdvCAM}~{\resizebox{0.9cm}{!}{\textnormal{\textcolor{gray}{[CVPR21]}}}}$ 
~\cite{lee2021anti} & 55.6 & 62.1 &  68.0  \\
$\textnormal{CLIMS}~{\resizebox{0.9cm}{!}{\textnormal{\textcolor{gray}{[CVPR22]}}}}$ 
~\cite{xie2022clims}& 56.6 & - & 70.5 \\
$\textnormal{SIPE}~{\resizebox{0.9cm}{!}{\textnormal{\textcolor{gray}{[CVPR22]}}}}$ 
~\cite{chen2022self} & 58.6 & 64.7 & 68.0 \\ 
$\textnormal{ESOL}~{\resizebox{1.1cm}{!}{\textnormal{\textcolor{gray}{[NeurIPS22]}}}}$ 
~\cite{li2022expansion} & 53.6 & 61.4 & 68.7 \\ 
$\textnormal{AEFT}~{\resizebox{0.9cm}{!}{\textnormal{\textcolor{gray}{[ECCV22]}}}}$ 
~\cite{yoon2022adversarial} & 56.0 & 63.5 & 71.0\\ 
$\textnormal{PPC}~{\resizebox{0.9cm}{!}{\textnormal{\textcolor{gray}{[CVPR22]}}}}$ 
~\cite{du2022weakly} & 61.5 & 64.0 & 64.0\\ 
$\textnormal{ReCAM}~{\resizebox{0.9cm}{!}{\textnormal{\textcolor{gray}{[CVPR22]}}}}$ 
~\cite{chen2022class} & 54.8 & 60.4 & 69.7\\
$\textnormal{Mat-Label}~{\resizebox{0.9cm}{!}{\textnormal{\textcolor{gray}{[ICCV23]}}}}$ 
~\cite{wang2023treating} & 62.3 & 65.8 & 72.9\\ 
$\textnormal{FPR}~{\resizebox{0.9cm}{!}{\textnormal{\textcolor{gray}{[ICCV23]}}}}$ 
~\cite{chen2023fpr} & 63.8 & 66.4 & 68.5\\ 
$\textnormal{LPCAM}~{\resizebox{0.9cm}{!}{\textnormal{\textcolor{gray}{[CVPR23]}}}}$ 
~\cite{chen2023extracting} & 62.1 & - & 72.2\\ 
$\textnormal{ACR}~{\resizebox{0.9cm}{!}{\textnormal{\textcolor{gray}{[CVPR23]}}}}$ 
~\cite{kweon2023weakly} & 60.3 & 65.9 & 72.3\\ 
$\textnormal{SFC}~{\resizebox{0.9cm}{!}{\textnormal{\textcolor{gray}{[AAAI24]}}}}$
~\cite{zhao2024sfc} & 64.7 & 69.4 & 73.7\\ 
\midrule
$\textnormal{IRN}~{\resizebox{0.9cm}{!}{\textnormal{\textcolor{gray}{[CVPR19]}}}}$ 
\cite{ahn2019weakly} & {48.8} & 53.7  & {66.5}\\ \rowcolor{gray!30}
{+CPAL (Ours)} &\textbf{62.5} \textcolor{blue!60}{$\uparrow$13.7} & \textbf{66.2} \textcolor{blue!60}{$\uparrow$12.5} & \textbf{72.7} \textcolor{blue!60}{$\uparrow$6.2}\\
\midrule
$\textnormal{AMN}~{\resizebox{0.9cm}{!}{\textnormal{\textcolor{gray}{[CVPR22]}}}}$  
~\cite{lee2022threshold} & 62.1 & 66.1 & 72.2\\ \rowcolor{gray!30}
+CPAL (Ours) &\textbf{65.7} \textcolor{blue!60}{$\uparrow$3.6}& \textbf{68.2} \textcolor{blue!60}{$\uparrow$2.1}  & \textbf{74.1} \textcolor{blue!60}{$\uparrow$1.9} \\
\midrule
$\textnormal{MCTformer}~{\resizebox{0.9cm}{!}{\textnormal{\textcolor{gray}{[CVPR22]}}}}$ 
~\cite{xu2022multi} & 61.7 & 64.5 & 69.1\\  \rowcolor{gray!30}
+CPAL (Ours) &\textbf{66.8} \textcolor{blue!60}{$\uparrow$5.1} & \textbf{69.3} \textcolor{blue!60}{$\uparrow$4.8} & \textbf{74.7} \textcolor{blue!60}{$\uparrow$5.6} \\
\midrule
$\textnormal{CLIP-ES}~{\resizebox{0.9cm}{!}{\textnormal{\textcolor{gray}{[CVPR23]}}}}$ 
~\cite{lin2023clip} &70.8 & -  & 75.0 \\ \rowcolor{gray!30}
+CPAL (Ours) &\textbf{71.9} \textcolor{blue!60}{$\uparrow$1.1} & - & \textbf{75.8} \textcolor{blue!60}{$\uparrow$0.8} \\
\bottomrule
\end{tabular}
\vspace{-0.25cm}
\label{labelVOC}
\end{table}

\subsection{Comparisons with State-of-the-Art Methods}

\noindent \textbf{Improved Localization Maps:} 
Since the proposed CPAL does not modify the architecture of the CAM network, it simply integrates the CPAL branch as supervision into multiple methods. Table~\ref{labelVOC} presents the results of applying CPAL to various well-known methods (IRN~\cite{ahn2019weakly}, AMN~\cite{lee2022threshold}, MCTformer~\cite{xu2022multi}, and CLIP-ES~\cite{lin2023clip}) and show improvements in localization maps on VOC 2012. For instance, incorporating CPAL into AMN improves performance by 3.6\% in seed and 2.1\% in pseudo masks. When plugging CPAL into the CLIP-ES model, there is a 1.1\% gain in the seed. Fig.~\ref{fig31} visualizes the comparison with baseline AMN and MCTformer, showing that CPAL can effectively capture high-quality localization maps.

\noindent \textbf{Improved Segmentation Results:} Table~\ref{miou_results} shows the performance of the semantic segmentation model trained with pseudo-labels generated by our method. Pseudo-labels are utilized to train the DeepLabV2 segmentation model. Comparisons with related works. Our AMN+CPAL achieves state-of-the-art results on VOC (mIoU of 72.5\% on the validation set and 72.9\% on the test set). On the more challenging MS COCO dataset, our MCTformer+CPAL (with ResNet-38 as the backbone) outperforms the state-of-the-art result AMN and all related works based on ResNet-38. For CLIP-ES, CPAL improves performance (+1.4\% mIoU on the COCO val). These superior results on both datasets confirm the effectiveness of our CPAL, which accurately captures the semantic features and object structures.

\begin{table}[t]
\centering
\scriptsize
\caption{The mIoU results (\%) based on DeepLabV2 on PASCAL VOC and MS COCO. $\mathcal{I}$ denotes using image-level labels. $\mathcal{S}$ denotes using saliency maps. $\mathcal{L}$ denotes using Language supervision.}
\vspace{-0.25cm}
\begin{tabular}{rlccccl}
\toprule[1pt]
% \midrule  %multicolumn
& \multirow{2}{*}{{Methods}} & \multirow{2}{*}{{Sup.}} & \multicolumn{2}{c}{{VOC}} & \multicolumn{1}{c}{{COCO}} \\ \cmidrule(r){4-6}
& & & {Val} & {Test} & {Val} \\   
\toprule[1pt]
\multirow{3}{*}{\rotatebox[origin=c]{90}{{Trans.}}}
&$\textnormal{AFA}~{\resizebox{0.9cm}{!}{\textnormal{\textcolor{gray}{[CVPR22]}}}}$ 
~\cite{ru2022learning} & $\mathcal{I}$ & 66.0 & 66.3 & - \\ 
&$\textnormal{BECO}~{\resizebox{0.9cm}{!}{\textnormal{\textcolor{gray}{[CVPR23]}}}}$ 
~\cite{rong2023boundary} &$\mathcal{I}$  & 73.7 & 73.5 & 42.0 \\ 
&$\textnormal{ToCo}~{\resizebox{0.9cm}{!}{\textnormal{\textcolor{gray}{[CVPR23]}}}}$ 
\cite{ru2023token} & $\mathcal{I}$ & 71.1 & 72.2 & 42.3 \\ 
\toprule[1pt]
\multirow{5}{*}{\rotatebox[origin=c]{90}{{ResNet38}}} 
&$\textnormal{Spatial-BCE}~{\resizebox{0.9cm}{!}{\textnormal{\textcolor{gray}{[ECCV22]}}}}$
~\cite{wu2022adaptive} & $\mathcal{I}+\mathcal{S}$ & 70.0 & 71.3 & 35.2 \\
&$\textnormal{MCTformer}~{\resizebox{0.9cm}{!}{\textnormal{\textcolor{gray}{[CVPR22]}}}}$  \cite{xu2022multi} & $\mathcal{I}$ & 71.9 & 71.6 & 42.0 \\ 
&$\textnormal{USAGE}~{\resizebox{0.9cm}{!}{\textnormal{\textcolor{gray}{[ICCV23]}}}}$ 
~\cite{peng2023usage} & $\mathcal{I}$ & 71.9 & 72.8 & 42.7 \\
&$\textnormal{OCR}~{\resizebox{0.9cm}{!}{\textnormal{\textcolor{gray}{[CVPR23]}}}}$+SEAM
~\cite{cheng2023out} & $\mathcal{I}$ & 67.8 & 68.4 & 33.2 \\
&$\textnormal{ACR}~{\resizebox{0.9cm}{!}{\textnormal{\textcolor{gray}{[ICCV23]}}}}$ 
~\cite{sun2023all} & $\mathcal{I}$ &71.9 &71.9 & 45.3 \\
&\cellcolor{gray!30} MCTformer +CPAL (Ours) & \cellcolor{gray!30} $\mathcal{I}$ & \cellcolor{gray!30} \textbf{72.8} &\cellcolor{gray!30}  \textbf{73.5} & \cellcolor{gray!30}  \textbf{46.5} \\ 
\toprule[1pt]
\multirow{17}{*}{\rotatebox[origin=c]{90}{{ResNet-101}}}
&$\textnormal{EPS}~{\resizebox{0.9cm}{!}{\textnormal{\textcolor{gray}{[CVPR21]}}}}$ 
~\cite{lee2021railroad} &$\mathcal{I}+\mathcal{S}$& 70.9 & 71.0 & - \\
&$\textnormal{RIB}~{\resizebox{1.1cm}{!}{\textnormal{\textcolor{gray}{[NeurIPS21]}}}}$ 
~\cite{lee2021reducing} & $\mathcal{I}+\mathcal{S}$ & 68.3 & 68.6 & 44.2 \\
&$\textnormal{EDAM}~{\resizebox{0.9cm}{!}{\textnormal{\textcolor{gray}{[CVPR21]}}}}$ 
~\cite{wu2021embedded} & $\mathcal{I}+\mathcal{S}$ & 70.9 & 71.8 & - \\ 
&$\textnormal{ESOL}~{\resizebox{1.1cm}{!}{\textnormal{\textcolor{gray}{[NeurIPS22]}}}}$ 
~\cite{li2022expansion} & $\mathcal{I}+\mathcal{S}$ & 69.9 & 69.3 & 42.6 \\
&$\textnormal{RCA}~{\resizebox{0.9cm}{!}{\textnormal{\textcolor{gray}{[CVPR22]}}}}$ 
~\cite{zhou2022regional} +OOA & $\mathcal{I}+\mathcal{S}$ & 72.2  & 72.8 & 36.8 \\
& $\textnormal{IRN}~{\resizebox{0.9cm}{!}{\textnormal{\textcolor{gray}{[CVPR19]}}}}$ 
~\cite{ahn2019weakly} & $\mathcal{I}$ & 63.5 & 64.8 & 42.0 \\
&$\textnormal{SEAM}~{\resizebox{0.9cm}{!}{\textnormal{\textcolor{gray}{[CVPR20]}}}}$ 
~\cite{wang2020self} & $\mathcal{I}$ & 64.5 & 65.7 & 32.8 \\ 
&$\textnormal{ReCAM}~{\resizebox{0.9cm}{!}{\textnormal{\textcolor{gray}{[CVPR22]}}}}$ 
~\cite{chen2022class} & $\mathcal{I}$ & 68.5 & 68.4 & 42.9 \\ 
&$\textnormal{OOD}~{\resizebox{0.9cm}{!}{\textnormal{\textcolor{gray}{[CVPR22]}}}}$ 
~\cite{lee2022weakly} +Adv & $\mathcal{I}$ & 69.8 & 69.9 & - \\
&$\textnormal{AMN}~{\resizebox{0.9cm}{!}{\textnormal{\textcolor{gray}{[CVPR22]}}}}$ 
~\cite{lee2022threshold} & $\mathcal{I}$ & 69.5 & 69.6 & 44.7 \\ 
&$\textnormal{SIPE}~{\resizebox{0.9cm}{!}{\textnormal{\textcolor{gray}{[CVPR22]}}}}$ 
~\cite{chen2022self} & $\mathcal{I}$ & 68.8 & 69.7 & 40.6 \\
&$\textnormal{LPCAM}~{\resizebox{0.9cm}{!}{\textnormal{\textcolor{gray}{[CVPR23]}}}}$ 
~\cite{chen2023extracting} +AMN & $\mathcal{I}$ & 70.1 & 70.4 & 45.5 \\
&$\textnormal{CLIMS}~{\resizebox{0.9cm}{!}{\textnormal{\textcolor{gray}{[CVPR22]}}}}$ 
\cite{xie2022clims} & $\mathcal{I+L}$ & 70.4 & 70.0 & - \\ 
&$\textnormal{CLIP-ES}~{\resizebox{0.9cm}{!}{\textnormal{\textcolor{gray}{[CVPR23]}}}}$ 
\cite{lin2023clip} & $\mathcal{I+L}$ & 73.8 & 73.9 & 45.4 \\ 
& \cellcolor{gray!30} IRN +CPAL (Ours) & \cellcolor{gray!30} $\mathcal{I}$ &  \cellcolor{gray!30} \textbf{71.8} &  \cellcolor{gray!30} \textbf{72.1} & \cellcolor{gray!30} \textbf{42.9} \\ 
& \cellcolor{gray!30} AMN +CPAL (Ours) & \cellcolor{gray!30} $\mathcal{I}$ &  \cellcolor{gray!30} \textbf{72.5} &  \cellcolor{gray!30} \textbf{72.9} & \cellcolor{gray!30}  \textbf{46.3} \\
&\cellcolor{gray!30} CLIP-ES +CPAL (Ours)& \cellcolor{gray!30}  $\mathcal{I+L}$ & \cellcolor{gray!30} \textbf{74.5} &\cellcolor{gray!30} \textbf{74.7}& \cellcolor{gray!30}  \textbf{46.8} \\ 
\toprule[1pt]
\end{tabular}
\label{miou_results}
\vspace{-0.4cm}
\end{table}

\section{Conclusion}
\quad In this work, we propose a novel context prototype-aware learning (CPAL) strategy for WSSS methods, which aims to alleviate the knowledge bias between instances and contexts. This method mines effective feature attributes in context clusters and adaptively selects and adjusts context prototypes to enhance representation capabilities. The core of our method is prototype awareness, which is achieved by context-aware prototypes to accurately capture the intra-class variation and feature distribution alignment. Extensive experiments under various settings show that the proposed method outperforms existing state-of-the-art methods, and ablation studies reveal the effectiveness of our CPAL.

{
    \small
    \bibliographystyle{ieeenat_fullname}
    \bibliography{main}
}


\end{document}